%% file: main.tex
\documentclass[10pt,twocolumn,letterpaper]{article}

\usepackage{cvpr}              % To produce the CAMERA-READY version
\input{preamble}

\definecolor{cvprblue}{rgb}{0.21,0.49,0.74}
\usepackage[pagebackref,breaklinks,colorlinks,allcolors=cvprblue]{hyperref}
\usepackage{subcaption}
\usepackage{multirow}
\usepackage[table,xcdraw]{xcolor}

\def\paperID{8101} % *** Enter the Paper ID here
\def\confName{CVPR}
\def\confYear{2026}

\newcommand{\ignore}[1]{}

\newcommand*{\affaddrMC}[1]{#1} % No op here. Customize it for different styles.
\newcommand*{\affmarkMC}[1][*]{\textsuperscript{#1}}

\title{\texttt{wrivinder}: Towards Spatial Intelligence for Geo-locating Ground Images onto Satellite Imagery}
\author{%
Chandrakanth Gudavalli\affmarkMC[1], Tajuddin Manhar Mohammed\affmarkMC[1], Abhay Yadav\affmarkMC[2],  \\ Ananth Vishnu Bhaskar \affmarkMC[1], Hardik Prajapati \affmarkMC[1], Cheng Peng \affmarkMC[2], \\ Rama Chellappa\affmarkMC[2], Shivkumar Chandrasekaran \affmarkMC[1], B. S. Manjunath\affmarkMC[1]\\
\\
\affaddrMC{\affmarkMC[1]Mayachitra, Inc.} \;
\affaddrMC{\affmarkMC[2]Johns Hopkins University} %\\
}
\begin{document}

\maketitle
\input{sec/0__abstract}
\input{sec/1__intro}

\input{sec/2__related_work}
\input{sec/3__dataset}

\input{sec/4__methodology}

\input{sec/exp_portrait}
\input{sec/acknowledgements}

% \input{main_suppl}

{
    \small
    \renewcommand{\refname}{Bibliography}
    \bibliographystyle{ieeenat_fullname}
    \bibliography{main}
}

\end{document}

%% file: preamble.tex
\usepackage[sectionbib]{chapterbib}

\usepackage{float}     % [H]
\usepackage{placeins}  % \FloatBarrier
\usepackage{graphicx}
\usepackage{subcaption} % for side-by-side panels
\usepackage{booktabs}
\usepackage{tabularx}
\usepackage{array}
\usepackage{makecell}

%% file: sec/0__abstract.tex
\begin{abstract}
Aligning ground-level imagery with geo-registered satellite maps is crucial for mapping, navigation, and situational awareness, yet remains challenging under large viewpoint gaps or when GPS is unreliable. We introduce \textbf{Wrivinder}, a zero-shot, geometry-driven framework that aggregates multiple ground photographs to reconstruct a consistent 3D scene and align it with overhead satellite imagery. Wrivinder combines SfM reconstruction, 3D Gaussian Splatting, semantic grounding, and monocular depth–based metric cues to produce a stable zenith-view rendering that can be directly matched to satellite context for metrically accurate camera geo-localization. To support systematic evaluation of this task—which lacks suitable benchmarks—we also release \textbf{MC-Sat}, a curated dataset linking multi-view ground imagery with geo-registered satellite tiles across diverse outdoor environments. Together, Wrivinder and MC-Sat provide a first comprehensive baseline and testbed for studying geometry-centered cross-view alignment without paired supervision. In zero-shot experiments, Wrivinder achieves sub-30\,m geolocation accuracy across both dense and large-area scenes, highlighting the promise of geometry-based aggregation for robust ground-to-satellite localization. The MC-Sat dataset and Wrivinder codebase will be publicly released.~\footnotemark
\end{abstract}
% \footnotetext{This paper is currently under review. MC-Sat Dataset along with a github will be publicly released upon completion of the review process.}
\footnotetext{Under review. The MC-Sat dataset and related resources will be released after the review process.}

% \vspace{-0.3cm}

\begin{figure}[!ht]
    \centering
    % \includegraphics[width=\columnwidth]{figures/sat_aligner_intro__condensed.pdf}
    % \caption{Overview of satellite-to-ground image alignment challenge. \textbf{Top Left:} Images coming from the front side of a building. \textbf{Top Right:} Images coming from the back side of the building. Geolocating these images onto the satellite view (bottom) is challenging due to severe perspective distortions.}
    % \includegraphics[width=0.9\columnwidth]{figures/sat_aligner_intro.pdf}
    % \caption{Overview of satellite-to-ground image alignment challenge. \textbf{Top:} Images coming from the front side of a building. \textbf{Bottom:} Images coming from the back side of the building. Geolocating these images onto the satellite view (middle) is challenging due to severe perspective distortions.}
    \includegraphics[width=\columnwidth]{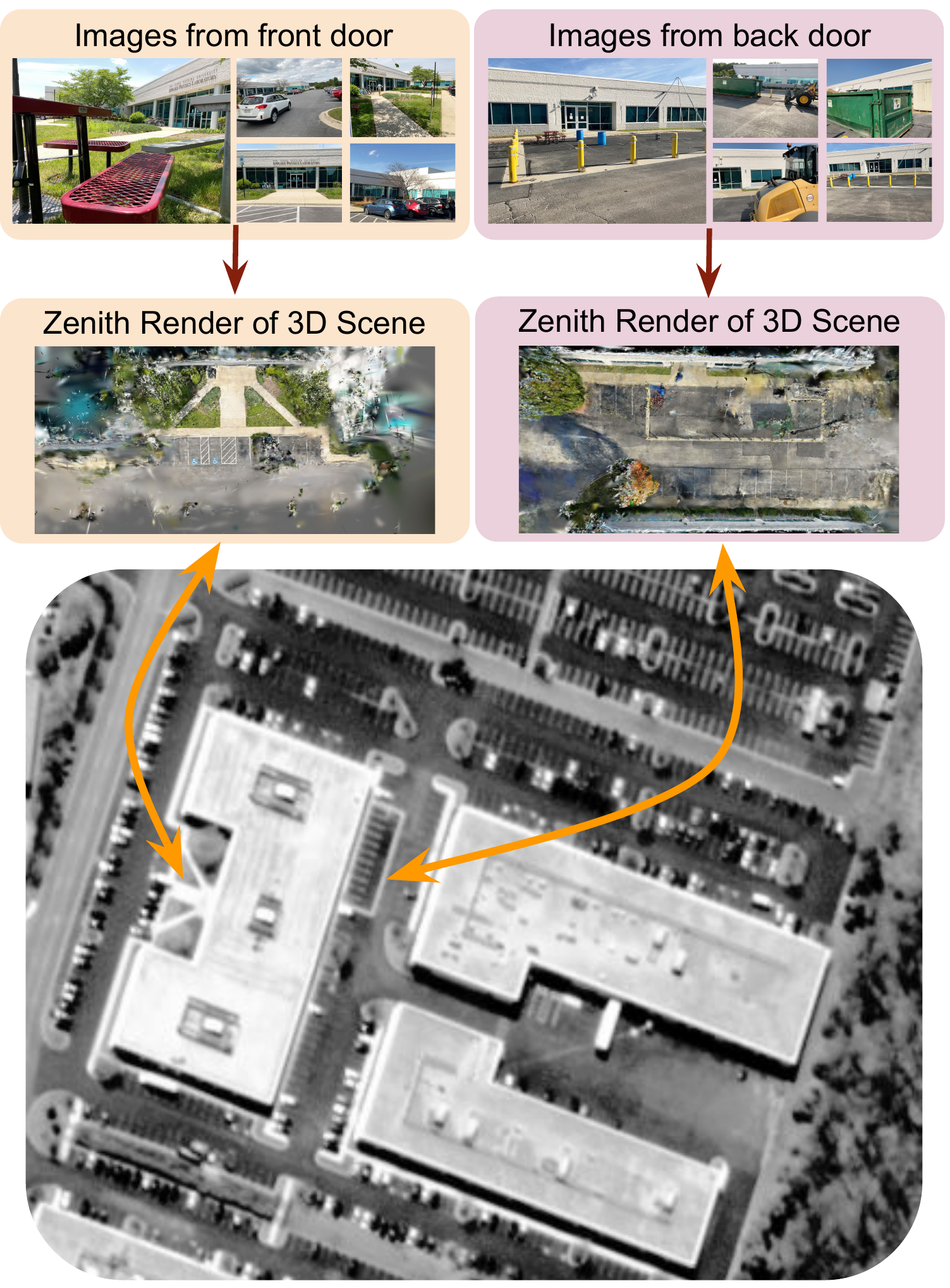}
    \caption{Overview of the satellite-to-ground image alignment pipeline. Directly aligning ground images to satellite views is impractical due to large viewpoint and scale differences. Wrivinder aggregates information from multiple ground images to reconstruct a 3D scene, generates a zenith-view rendering, and aligns it to the satellite image using the estimated metric dimensions in meters.}
    % \caption{Overview of satellite-to-ground image alignment pipeline. Aligning the ground images directly onto satellite images is not practical. Wrivinder aggregates the information from multiple ground images into a zenith render which will later aligned to satellite image by considering the estimated metric dimensions in meters}
    % \caption{Overview of satellite-to-ground image alignment challenge. \textbf{Top:} Images coming from the front side of a building. \textbf{Bottom:} Images coming from the back side of the building. Geolocating these images onto the satellite view (middle) is challenging due to severe perspective distortions.}
    
    \label{fig:intro}
    % \vspace{-0.5cm}
\end{figure}

%% file: sec/1__intro.tex
\section{Introduction}
\label{sec:intro}

Accurately aligning ground-level imagery with geo-registered satellite maps is central to applications such as autonomous navigation~\cite{boroujeni2024comprehensive, zhu2021deep}, disaster response~\cite{cejudo2023emergency}, large-scale mapping~\cite{shan2014accurate, roros2023multi}, and situational awareness in GPS-denied environments~\cite{viswanathan2014vision}. Recovering camera GPS coordinates directly from unorganized ground photos and a corresponding satellite tile would enable consistent geo-referencing of ground imagery and support robust map construction pipelines without relying on GPS.

The task, however, remains extremely challenging due to the drastic viewpoint, scale, and appearance differences between ground and satellite imagery (Fig.~\ref{fig:intro}). The same region can look radically different across changes in altitude, orientation, and occlusion, resulting in perspective distortions and geometric ambiguities. Existing cross-view geo-localization (CVGL) approaches learn ground–overhead correspondences from large quantities of paired, geo-aligned data, achieving strong results on structured, road-centric benchmarks. Yet such paired supervision is scarce in real-world environments—campuses, construction sites, or rural regions—leading to poor generalization under distribution shift.

Most CVGL methods, including sequence-based (SeqGeo) and set-based (Set-CVGL) variants, treat the task as supervised retrieval: given a ground image, retrieve the most similar satellite crop. These models rarely operate in a true zero-shot regime and produce a nearest-neighbor satellite tile rather than a physically meaningful camera pose or GPS coordinate. Moreover, their 2D feature representations lack explicit 3D reasoning, limiting robustness to large viewpoint gaps. In contrast, \textbf{Wrivinder} departs from the retrieval paradigm by leveraging geometric reconstruction and metric alignment to infer physically grounded camera locations without paired training data.

To achieve this, Wrivinder aggregates geometric and semantic cues from multiple ground images as a bridge to the satellite domain. Given an unordered set of ground photos, it first reconstructs a sparse 3D scene using Structure-from-Motion (SfM). Monocular depth priors and semantic masks provide scene orientation and ground-plane estimates, enabling a consistent zenith viewpoint. A dense, photorealistic 3D Gaussian Splatting (3DGS) model is then rendered from this viewpoint. A test-time, self-supervised Deep Template Matcher aligns the zenith render to the geo-registered satellite tile, yielding pixel-level correspondences that are back-projected through the 3DGS and SfM models to estimate camera GPS positions. This geometry-centered formulation enables metrically meaningful, training-free localization across diverse and unconstrained environments.

\noindent \textbf{Main Contributions.}
\begin{itemize}
    \item \textbf{MC-Sat Dataset.} We curate and release \textbf{MC-Sat}, the first dataset linking multi-view ground imagery, SfM/3DGS reconstructions, and geo-registered satellite context across diverse outdoor environments. MC-Sat fills a critical gap in CVGL benchmarks by enabling metrically evaluated, multi-view, and truly zero-shot ground-to-satellite alignment in unconstrained scenes.

    \item \textbf{Wrivinder Framework.} We propose \textbf{Wrivinder}, a geometry-driven, zero-shot framework that reconstructs a consistent 3D scene from multiple ground images and aligns it with overhead satellite imagery. Wrivinder integrates SfM, 3DGS, semantic grounding, and metric depth cues to obtain physically meaningful camera GPS estimates without paired supervision.

    \item \textbf{Test-Time Self-Supervised Alignment.} We develop a lightweight, \textbf{test-time self-supervised} Deep Template Matcher that aligns zenith-view 3DGS renderings to satellite images, enabling robust cross-view correspondence under extreme viewpoint changes and without any ground–satellite training pairs.
\end{itemize}

%% file: sec/2__related_work.tex
\section{Related Work}
\label{sec:rel_work}
\paragraph{Cross-View Geo-Localization (CVGL).}
The dominant paradigm in CVGL relies on supervised learning from paired ground–satellite images. Benchmarks such as CVUSA~\cite{workman2015wide} and CVACT~\cite{liu2019lending} have enabled models based on Siamese CNNs~\cite{hu2018cvm}, spatial-aware attention~\cite{shi2019spatial}, and transformers~\cite{zhu2022transgeo,ding2022layer}. Recent methods reach near-saturated performance—e.g., Sample4Geo~\cite{deuser2023sample4geo} achieves 97.83\% Recall@1 on CVUSA—while Set-CVGL~\cite{wu2024cross} extends retrieval to multi-view inputs. Sequence-based methods~\cite{zhang2023cross,pillai2024garet} further incorporate temporal context. However, all of these approaches remain tightly coupled to curated, road-centric benchmarks and require large quantities of paired, geo-aligned data. They generalize poorly to unconstrained scenes (e.g., campuses, construction zones, or rural landscapes), and even unsupervised~\cite{li2024first} or weakly-supervised~\cite{shi2024fine} variants still depend on dataset-specific adaptation. In contrast, \textbf{Wrivinder} operates in a genuinely zero-shot setting—requiring no paired supervision, no fine-tuning, and no dataset-specific training.

% \subsection*{Geometric Methods} 
% \paragraph{Geometric Methods.}
\paragraph{Geometry-Based Alignment.}
Before deep learning, geometric methods aligned ground imagery to overhead views via SfM and handcrafted cost functions. Kaminsky et al.~\cite{kaminsky2009alignment} matched sparse SfM reconstructions to satellite imagery using edge and free-space cues. While effective, sparse point clouds lack photorealistic appearance and are difficult to match under large viewpoint gaps. \textbf{Wrivinder} advances this classical geometric lineage by replacing sparse SfM points with dense, appearance-preserving 3D Gaussian Splatting (3DGS), enabling robust photometric alignment through realistic zenith-view rendering.

\begin{figure*}[!ht]
    \centering
    \includegraphics[width=2.0\columnwidth]{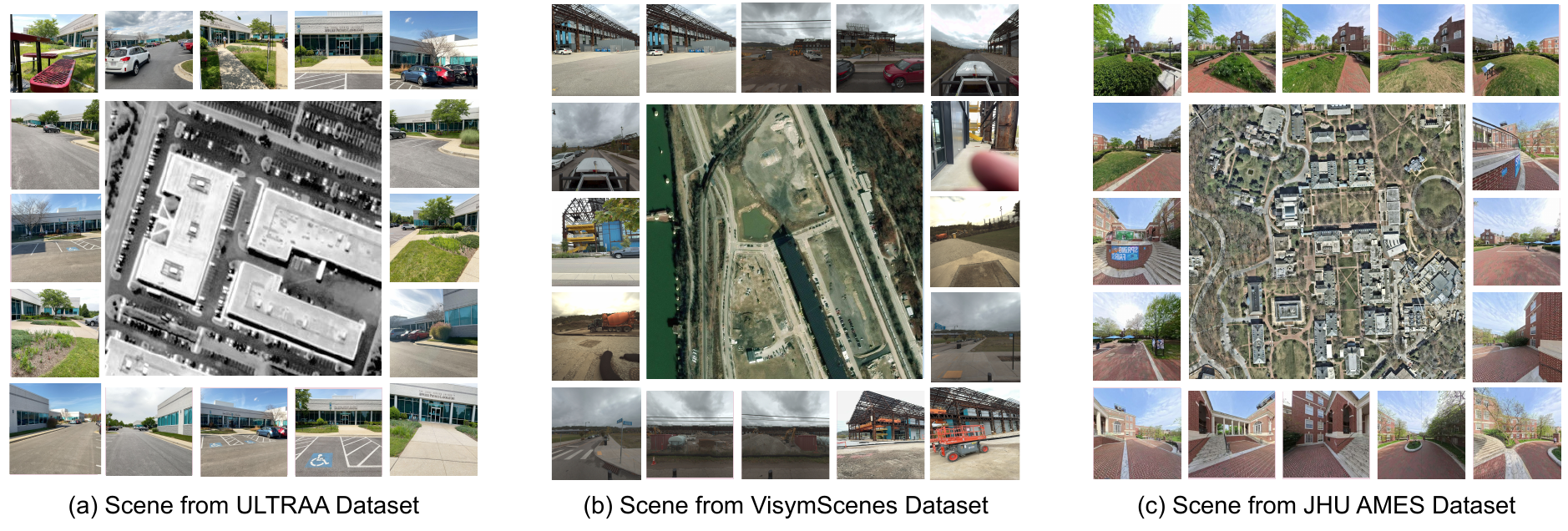}
    % \vspace{-0.4cm}
    \caption{MC-Sat Dataset Overview, showing scenes from the \textit{ULTRAA}, \textit{VisymScenes}, and \textit{JHU-Ames} datasets. The central image in each tile is a satellite view of the scene, surrounded by corresponding ground images illustrating the diversity of viewpoints and environments.}
    \vspace{-0.3cm}
    \label{fig:datasets_vis}
\end{figure*}

% \paragraph{Neural Rendering Approaches.}
%Neural Radiance Fields have been applied to cross-view tasks. Bredvik et al.~\cite{bredvik2025metadata} use NeRF to render nadir views from airborne imagery for satellite alignment, achieving 0.83-6.03m accuracy. Sat-NeRF~\cite{mari2022satnerf} and its follow-ups~\cite{11224549, Mari_2023_CVPR} addresses multi-view satellite photogrammetry. 
% While effective, NeRF requires 8-24 hours training and renders at 1-5 FPS. Wrivinder adopts 3D Gaussian Splatting for 10-100$\times$ faster training, 90-150+ FPS rendering, and superior view synthesis quality for perspectives similar to training views—precisely our zenith rendering scenario. 
%While effective, traditional NeRF suffers from slow training and low frame rates. Wrivinder uses 3D Gaussian Splatting for faster training and real-time rendering speeds, and superior view synthesis quality. Recent works~\cite{li2024geomgs,wang2024splatloc,wu2024gsplatloc} also demonstrate 3DGS's geometric accuracy advantages.

\paragraph{Neural Rendering and Photogrammetry.}
Neural radiance fields (NeRFs) have been explored for cross-view and photogrammetric tasks, including nadir-view synthesis for satellite alignment~\cite{bredvik2025metadata} and satellite multi-view reconstruction~\cite{mari2022satnerf,11224549,Mari_2023_CVPR}. However, conventional NeRFs are computationally intensive and slow to train. In contrast, \textbf{Wrivinder} leverages 3DGS for real-time rendering, fast convergence, and high-fidelity novel views, while maintaining the geometric precision of classical SfM. Recent advances in 3DGS~\cite{li2024geomgs,wang2024splatloc,wu2024gsplatloc} further motivate its use as a practical representation for zero-shot alignment.

% \paragraph{Learned View Transformations.}
%Other approaches employ learned transformations. Bird's-eye-view methods~\cite{wang2024cross} learn neural transformations from ground to overhead views, achieving strong results through supervised training. However, these require paired training data and assume ground-plane geometry. Wrivinder performs explicit 3D reconstruction, where zenith rendering is direct geometric projection without learned transformations, enabling zero-shot deployment and handling arbitrary 3D structure. Foundation models~\cite{chen2024dinov2,wu2025crosstext2loc} offer complementary capabilities but still require task-specific adaptation.

%By combining SfM reconstruction, 3DGS neural rendering, and geometric zenith-satellite alignment, Wrivinder uniquely achieves zero-shot cross-view localization without training, paired data, or geometric assumptions.

\paragraph{Learning Ground-to-Overhead Mappings.}
Other approaches attempt to learn transformations from ground views to top-down representations. BEV-based models~\cite{wang2024cross} achieve strong performance on benchmark datasets but rely on paired supervision and often assume a flat ground plane. Foundation models~\cite{chen2024dinov2,wu2025crosstext2loc} provide strong semantic priors yet still require task-specific adaptation for cross-view alignment. \textbf{Wrivinder}, by contrast, performs explicit 3D reconstruction: its zenith-view render arises from geometric projection rather than learned mapping, allowing true zero-shot deployment and accommodating complex 3D structure.

\paragraph{Summary.}
By integrating SfM-based reconstruction, 3DGS neural rendering, and geometric zenith–satellite alignment, \textbf{Wrivinder} unifies the interpretability of geometric methods with the realism of neural representations, achieving zero-shot cross-view localization without training, paired data, or restrictive planar assumptions.
%\textit{To further advance this direction, we introduce the MC-Sat dataset, which provides the first unified benchmark supporting geometry-driven and zero-shot cross-view localization research.}

%% file: sec/3__dataset.tex
\section{MC-Sat Dataset}
\label{sec:dataset}

To advance research in cross-view alignment beyond road-centric and paired-image benchmarks, we introduce the \textbf{MC-Sat} (Multi-view Capture–Satellite) dataset. MC-Sat is the first unified benchmark that jointly links multi-view ground imagery, 3D reconstructions, and geo-registered satellite context across diverse outdoor environments. It combines high-resolution overhead imagery with heterogeneous ground captures (Fig.~\ref{fig:datasets_vis}), enabling rigorous evaluation of geometry-based and zero-shot methods for satellite-to-ground alignment at metric precision. By providing aligned ground, 3D, and satellite views for unconstrained scenes—where paired supervision is typically unavailable—MC-Sat fills a critical gap in current CVGL datasets.

\subsection*{Dataset Construction}

\textbf{MC-Sat} integrates multiple complementary ground-image sources to capture broad geographic and geometric diversity. We aggregate multi-view imagery from ULTRAA~\cite{2zs6-ht63-24}, VisymScenes~\cite{xiangli2025doppelgangers++}, ACC-NVS~\cite{sugg2025accenture}, and JHU-Ames~\cite{li2025msgs}, spanning a wide range of environmental conditions, sensor types, and capture geometries (Table~\ref{tab:datasets}). For each site, geo-registered satellite or aerial imagery is obtained from the USDA NAIP program and the ESRI World Imagery basemap. Satellite tiles are selected to overlap the ground-image footprints using available geospatial metadata, producing consistent ground–satellite associations suitable for evaluating zero-shot, geometry-centered localization methods.

\subsection*{Ground Image Sources}

\textbf{ULTRAA}~\cite{2zs6-ht63-24} benchmarks view synthesis under sparse, heterogeneous captures with mixed camera intrinsics. It provides three challenging scenes from the Johns Hopkins APL and the Muscatatuck Urban Training Center (MUTC), offering varied geometry and limited-view overlap to test reconstruction robustness.

\textbf{VisymScenes}~\cite{xiangli2025doppelgangers++} contains 258K images from 149 sites across 42 cities and 15 countries. Each frame includes GPS, IMU, and intrinsic metadata, contributing substantial geographic, environmental, and sensor diversity, including realistic noise conditions.

\textbf{ACC-NVS1}~\cite{sugg2025accenture} includes 148K ground and airborne images across six scenes in Austin and Pittsburgh. Its multi-altitude captures under varying illumination and weather enrich MC-Sat with additional domain diversity and support evaluation in dynamic outdoor settings.

\textbf{JHU-Ames}~\cite{li2025msgs} offers 1.7K images of a single outdoor campus scene, providing a controlled setting for studying geometric and photometric consistency. Since only a subset includes absolute GPS, we align the remaining frames via relative SfM poses to ensure a consistent world coordinate frame.

\subsubsection*{Satellite Component (NAIP and ESRI)}

The overhead imagery in \textbf{MC-Sat} is drawn from two complementary sources: the USDA \textit{National Agriculture Imagery Program (NAIP)}~\cite{usgs_naip_2018} and the \textit{ESRI World Imagery} basemap~\cite{zeiler1999modeling}. NAIP provides orthorectified aerial imagery at 0.6–1.0~m/pixel resolution across the United States, offering high-fidelity detail suitable for metric-scale evaluation. ESRI World Imagery supplies globally consistent, geo-referenced overhead coverage, allowing inclusion of international scenes. For each MC-Sat site, satellite tiles are selected based on footprint overlap and image quality, aligned using available geospatial metadata, and manually checked for geometric consistency with the reconstructed 3D scenes.

\subsubsection*{Scale, Coverage, and Intended Use}

The released \textbf{MC-Sat} dataset comprises \textit{15 multi-view scenes} drawn from ULTRAA, VisymScenes, ACC-NVS, and JHU-Ames, totaling roughly \textit{20K ground images}. Each scene includes geo-registered satellite imagery from \textit{NAIP} or \textit{ESRI}, aligned to the reconstructed ground-image footprints. By linking dense multi-view captures with high-resolution overhead context, MC-Sat enables quantitative evaluation of geometry-centered and zero-shot localization pipelines, including \textit{Wrivinder}. Table~\ref{tab:quant_res} summarizes all scenes and associated metrics; additional statistics and visualizations are provided in the supplementary material.

MC-Sat includes two types of scenes: \textit{Image Density} and \textit{Reconstructed Area}. Image Density scenes feature many images concentrated around a small region (e.g., a building entrance or courtyard), supporting evaluation of fine-grained geometric alignment. Reconstructed Area scenes span larger spatial extents—building clusters or campus-scale environments—enabling assessment of long-range 3D reconstruction and satellite alignment. Together, these categories offer complementary settings for studying both local and global geo-localization performance in unconstrained outdoor environments.

% \paragraph{Summary and Transition.}  
In summary, the \textbf{MC-Sat} dataset provides the empirical foundation for evaluating geometry-driven, zero-shot cross-view localization under realistic, unconstrained outdoor conditions. Its diverse ground–satellite pairs, spanning both localized and large-scale scenes, enable systematic analysis of reconstruction accuracy, alignment precision, and generalization beyond road-centric benchmarks. We next describe \textbf{Wrivinder}, our proposed framework that uses these multi-view scenes to reconstruct consistent 3D geometry, render zenith-view representations, and align them to satellite imagery for metrically accurate camera geo-localization.

%% file: sec/4__methodology.tex
\begin{table}[!ht]
\centering
\small
\setlength{\tabcolsep}{4pt}

\begin{tabular}{lccc}
\hline
% \rowcolor[HTML]{F9CB9C}
\textbf{Dataset} &
\textbf{\#Scenes} &
\textbf{\#Images} &
\textbf{Imagery Type} \\
\hline
\textit{ULTRAA~\cite{2zs6-ht63-24}} & 3 & 1,028 & Ground \\
\textit{VisymScenes~\cite{xiangli2025doppelgangers++}} & 149 & 258K & Ground \\
\textit{ACC\mbox{-}NVS1~\cite{sugg2025accenture}} & 6 & 148K & Ground + Airborne \\
\textit{JHU-Ames~\cite{li2025msgs}} & 1 & 1,717 & Ground + Airborne \\
\hline
\end{tabular}

\vspace{-0.6em}
\caption{Overview of ground imagery datasets incorporated into the MC-Sat dataset. 
MC-Sat integrates subsets of these sources and pairs them with orthorectified satellite imagery. 
More details about the curated subset are reported in Table~\ref{tab:quant_res}.}
\label{tab:datasets}
\vspace{-0.6cm}
\normalsize
\end{table}

\begin{figure*}[!htbp]
    \centering
    \includegraphics[width=2.0\columnwidth]{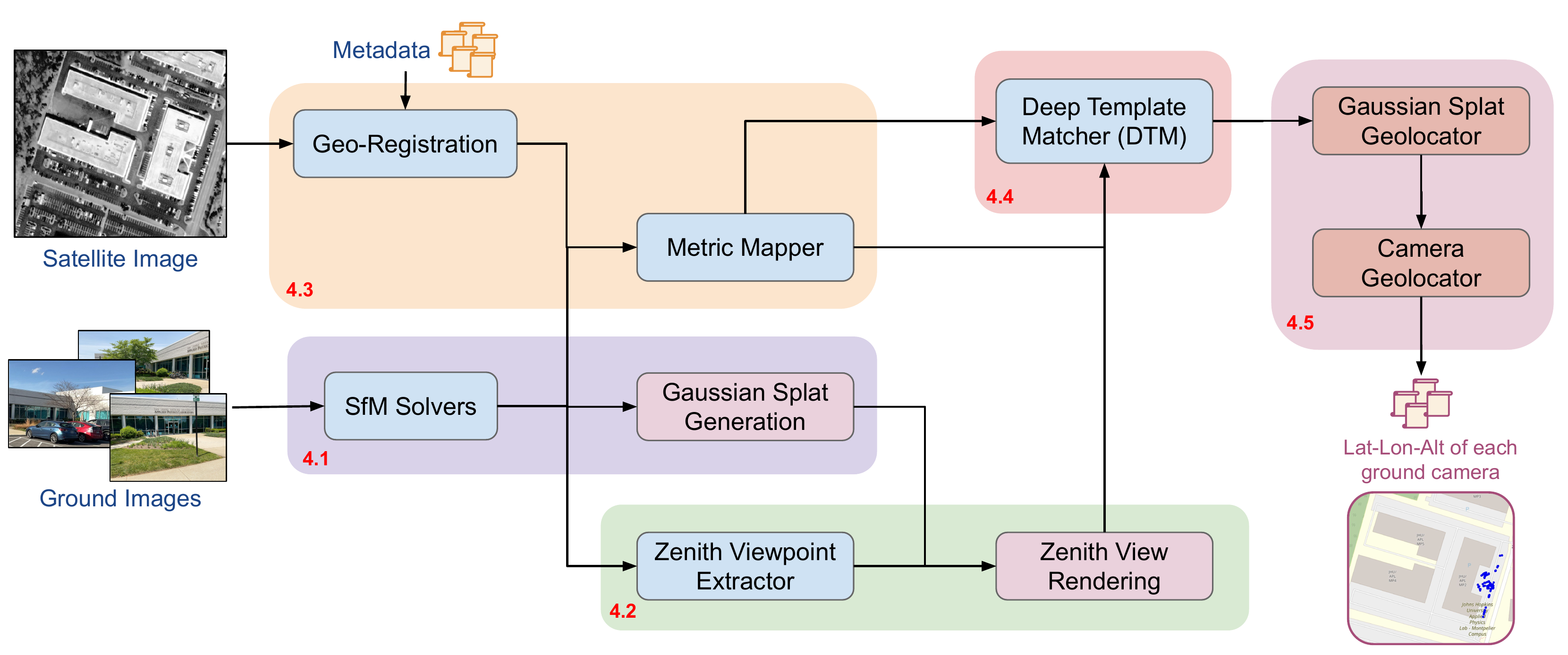}
    \vspace{-0.4cm}
    % \caption{Methodology Figure.}
    % \caption{Overview of Wrivinder, zero shot training free pipeline for geolocating ground images onto satellite image. Ground images are first processed through \textit{SfM solvers} to reconstruct a 3D point cloud, which is further refined by \textit{3D Gaussian Splatting}. The \textit{Zenith Viewpoint Extractor} determines the upward vertical and produces a \textit{Zenith View Rendering}. The \textit{Metric Mapper} estimates real-world dimensions of dimensions of zenith render using depth priors. \textit{Deep Template Matcher (DTM)} aligns the zenith render with the satellite image, after which the \textit{Gaussian Splat Geolocator} and \textit{Camera Geolocator} recover the GPS coordinates (lat–lon–alt) of all ground cameras.}
    \caption{
Overview of \textbf{Wrivinder}, a zero-shot, training-free pipeline for geo-locating ground images on a geo-registered satellite map. Given an unordered set of ground images, the pipeline reconstructs a sparse 3D scene via \textit{SfM} and densifies it using \textit{3D Gaussian Splatting}. The \textit{Zenith Viewpoint Extractor} estimates the vertical direction and generates a top-down zenith render. The \textit{Metric Mapper} uses monocular depth priors to recover approximate metric scale and determine the physical footprint of the zenith view. A test-time \textit{Deep Template Matcher (DTM)} aligns this render to the satellite image, and the resulting correspondences are back-projected through the 3DGS and SfM models via the \textit{Gaussian Splat Geolocator} to estimate GPS positions for all ground cameras.
}
    \vspace{-0.5cm}
    \label{fig:pipeline}
\end{figure*}

\section{Methodology}
\label{sec:method}

Given an unordered set of ground-level images and a corresponding geo-registered satellite view, \textbf{Wrivinder} aims to recover metrically accurate GPS locations for all ground cameras in a fully zero-shot setting. The key idea is to use \textit{geometry as the bridge} between drastically different viewpoints: instead of learning cross-domain correspondences, Wrivinder reconstructs a 3D representation of the scene and aligns it directly to the satellite frame through geometric projection and self-supervised matching.

The pipeline consists of five stages. We first reconstruct sparse scene geometry using a standard Structure-from-Motion (SfM) solver and densify it with a 3D Gaussian Splatting model (Sec.~\ref{sec:sfm3dgs}). We then estimate the vertical direction to generate a consistent zenith-view rendering (Sec.~\ref{sec:zenith}). Monocular depth cues provide approximate metric scale and determine the physical footprint of this zenith view (Sec.~\ref{sec:metric}). A lightweight, test-time self-supervised Deep Template Matcher aligns the zenith render to the satellite image (Sec.~\ref{sec:dtm}). The resulting correspondences are finally back-projected through the 3DGS and SfM models to estimate GPS coordinates for all ground cameras (Sec.~\ref{sec:geolocator}).

\subsection{3D Reconstruction (SfM + 3DGS)}
\label{sec:sfm3dgs}

We begin by reconstructing scene geometry using standard Structure-from-Motion (SfM) solvers such as HLOC+COLMAP~\cite{sarlin2019coarse}, GLOMAP~\cite{pan2024global}, or VGGT-style~\cite{wang2025vggt} pipelines. These methods estimate camera intrinsics, extrinsics, and a sparse 3D point cloud in an arbitrary relative coordinate frame.

To obtain a dense and photorealistic representation in the same coordinate system, we further refine the reconstruction using 3D Gaussian Splatting (3DGS) methods such as Scaffold-GS~\cite{lu2024scaffold} or Octree-GS~\cite{ren2024octree}. Unlike sparse SfM points, 3DGS jointly optimizes Gaussian primitives for both geometry and appearance, suppressing floating artifacts and producing high-fidelity renderings suitable for stable zenith-view synthesis.

% Each 3D point reconstructed by the SfM pipeline corresponds to at least one pixel in the input ground images. Using semantic segmentation maps of these images, we propagate the semantic labels (e.g., roads, walkways, vegetation, sky) onto the 3D points, allowing us to distinguish ground plane points and non-ground plane points. Assuming that most ground-level images are captured within approximately two meters of the ground surface, we estimate the ground plane by fitting a plane to the estimated camera locations and 3D points correspond to ground labels.
\subsection{Zenith Rendering of Ground Clusters}
\label{sec:zenith}

Each SfM point corresponds to at least one pixel in the input ground images. To identify ground-plane structure, we obtain semantic masks for all images using a Mask2Former model with a BEiTv2 Adapter backbone (large variant, 896$\times$896)~\cite{chen2022vision}, pretrained on COCO-Stuff~\cite{Caesar_2018_CVPR}. The model predicts 172 categories spanning both ``things'' and ``stuff.'' Pixel-level labels are propagated to the triangulated SfM points, enabling separation of ground surfaces from surrounding structures.

Ground-relevant classes include \textit{road, sidewalk, grass, dirt, gravel, pavement, ground-other, sand, playingfield}, along with context-dependent floor materials such as \textit{marble, stone, tile, wood, carpet, platform}, and \textit{bridge} surfaces. These categories reliably identify traversable ground regions. Assuming ground-level cameras are captured within roughly two meters of the ground plane, we estimate a consistent ground plane by jointly fitting a plane to (i) all SfM points with ground-like semantic labels and (ii) the recovered camera centers.

\vspace{-0.3cm}

\paragraph{Ground plane and vertical estimation.}
To determine a consistent top-down (zenith) viewpoint, we analyze the geometry of the sparse SfM point cloud. Let $\mathcal{P}=\{\mathbf{x}_i\}_{i=1}^N$ denote all 3D points. We compute the centroid
\vspace{-0.25cm}
\[
\mathbf{c} = \frac{1}{N}\sum_{i=1}^N \mathbf{x}_i,
\]
and apply PCA to the centered points $\mathbf{x}_i - \mathbf{c}$ via the covariance matrix
\vspace{-0.25cm}
\[
\mathbf{\Sigma} = \frac{1}{N} \sum_{i=1}^N 
(\mathbf{x}_i-\mathbf{c})(\mathbf{x}_i-\mathbf{c})^\top.
\]
Let $\mathbf{v}_1,\mathbf{v}_2,\mathbf{v}_3$ be eigenvectors of $\mathbf{\Sigma}$ in decreasing order of eigenvalues. The smallest-variance direction $\mathbf{v}_3$ typically aligns with the ground-plane normal in outdoor scenes, and we treat it as the vertical axis.

To resolve its sign ambiguity, we compare $\mathbf{v}_3$ with the mean camera center $\bar{\mathbf{c}}$. If most cameras lie in the negative half-space of $\mathbf{v}_3$, we flip the vector. The final vertical direction is
\vspace{-0.2cm}
\[
\hat{\mathbf{z}} = \operatorname{sign}\!\bigl((\bar{\mathbf{c}} - \mathbf{c})^\top \mathbf{v}_3\bigr)\mathbf{v}_3.
\]

\begin{figure}[!htbp]
    \centering
    \includegraphics[width=\columnwidth]{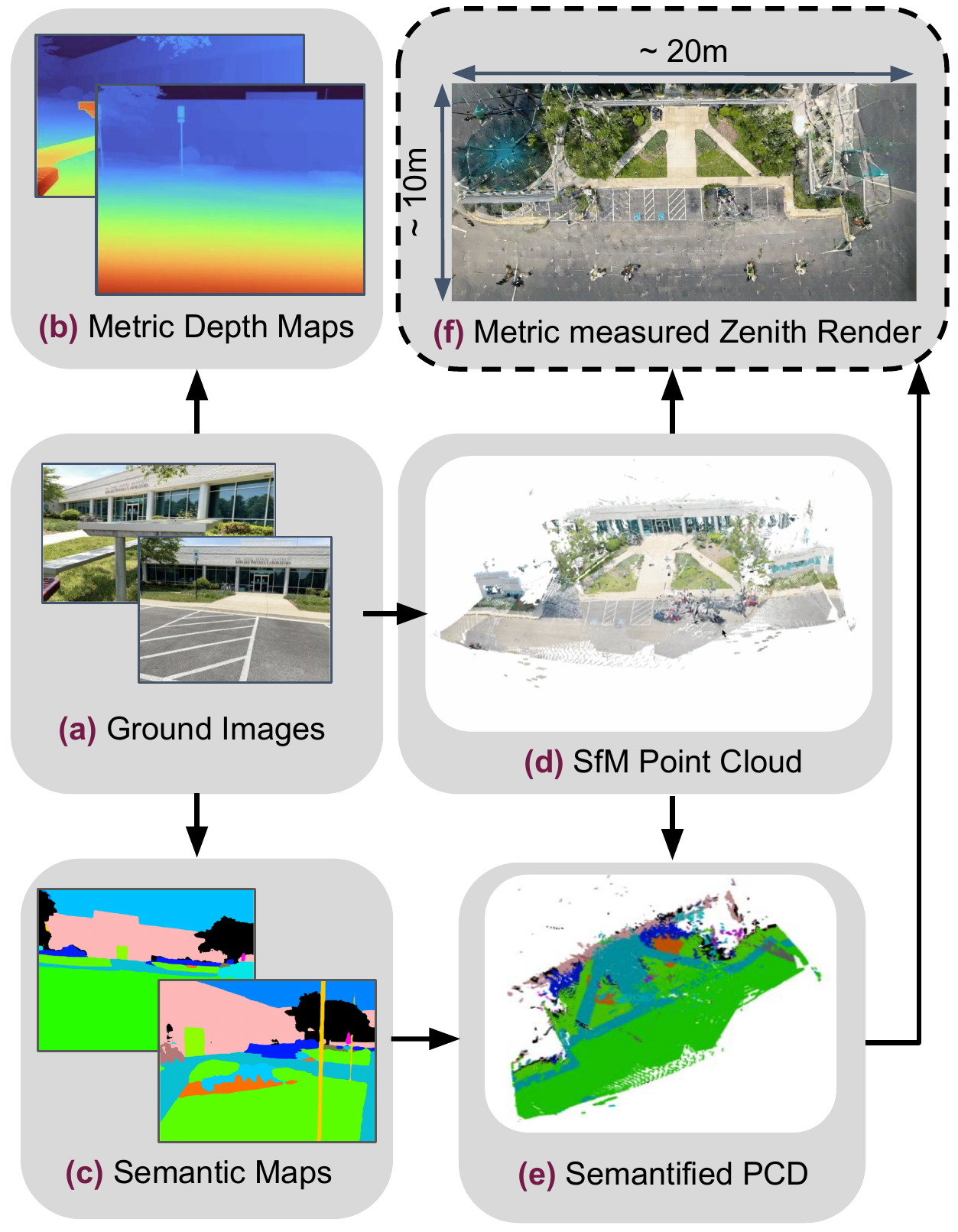}
    \caption{Key intermediate outputs of Wrivinder, showing semantic maps, the SfM point cloud, semantified reconstruction, metric depth maps, and the resulting metric-scaled zenith render.}
    \label{fig:blocks_wrivinder}
    \vspace{-0.5cm}
\end{figure}

\paragraph{Zenith viewpoint estimation.}
With the vertical axis $\hat{\mathbf{z}}$ determined, we construct an orthonormal basis for the zenith camera. We use $\mathbf{v}_1$ (the direction of maximum variance) as the in-plane axis $\hat{\mathbf{x}}$, and obtain the third axis via
\vspace{-0.25cm}
\[
\hat{\mathbf{x}}=\mathbf{v}_1,\qquad 
\hat{\mathbf{y}}=\hat{\mathbf{z}}\times\hat{\mathbf{x}},
\]
forming the rotation matrix
\vspace{-0.25cm}
\[
\mathbf{R}_{\text{zenith}} = 
\bigl[\,\hat{\mathbf{x}},\,\hat{\mathbf{y}},\,\hat{\mathbf{z}}\,\bigr]^\top.
\]

We place a virtual camera above the scene at
\vspace{-0.25cm}
\[
\mathbf{p} = \mathbf{c} + \delta\,\hat{\mathbf{z}},
\]
where $\delta$ is set from the robust spatial extent of the point cloud (98th-percentile radius in the PCA frame) to ensure full scene coverage. A standard look-at transformation from $\mathbf{p}$ toward $\mathbf{c}$ with up-vector $\hat{\mathbf{x}}$ yields a consistent zenith viewpoint.

Because the SfM and 3DGS reconstructions share the same coordinate system, this zenith camera can be applied directly to either representation, ensuring geometric consistency throughout the pipeline.

\subsection{Metric Mapper for Satellite Pixel Footprint}
\label{sec:metric}

We estimate approximate metric scale using monocular depth models such as \textit{DepthPro}~\cite{Bochkovskii2024} or \textit{PatchFusion}~\cite{li2024patchfusion}. Although noisy, predicted depths provide a consistent estimate of camera-to-pixel distance. For any image $i$ with pose $(\mathbf{R}_i,\mathbf{t}_i)$, the SfM depth of a 3D point $\mathbf{X}^{\text{sfm}}_k$ is  
\vspace{-0.2cm}
\[
z^{\text{sfm}}_k = \mathbf{e}_3^\top(\mathbf{R}_i \mathbf{X}^{\text{sfm}}_k + \mathbf{t}_i),
\]
while the predicted metric depth at the corresponding pixel is $d^{\text{pred}}_k = D_i(u_k,v_k)$.  
We assume a global scale $s$ relating SfM and metric depths,
\vspace{-0.2cm}
\[
d^{\text{pred}}_k \approx s\, z^{\text{sfm}}_k,
\]
and obtain an image-level estimate via least squares,
\[
s_i^\star = \frac{\sum_k z^{\text{sfm}}_k\, d^{\text{pred}}_k}{\sum_k (z^{\text{sfm}}_k)^2}.
\]

To mitigate outliers, we refine $s_i^\star$ using RANSAC over depth pairs $(z^{\text{sfm}}_k, d^{\text{pred}}_k)$ and select the scale with the highest inlier support. Among all images with valid tracks, the scale yielding the lowest reconstruction error is chosen as the global scale $\hat{s}$, which is applied uniformly:
\[
\mathbf{X}^{\text{metric}}_k = \hat{s}\,\mathbf{X}^{\text{sfm}}_k.
\]

Metric 3D points are then projected into the zenith coordinate frame, yielding coordinates $\mathbf{z}_k=(z_k^{(x)}, z_k^{(y)}, z_k^{(z)})$. A tight bounding box over the $(x,y)$ coordinates provides the physical footprint of the reconstruction:
\[
W_{\text{m}} = \max_k z^{(x)}_k - \min_k z^{(x)}_k,\qquad
H_{\text{m}} = \max_k z^{(y)}_k - \min_k z^{(y)}_k.
\]

Given the satellite’s ground sampling distance $g$ (meters/pixel), these extents convert to expected pixel dimensions,
\[
W_{\text{px}} \approx \tfrac{W_{\text{m}}}{g}, \qquad
H_{\text{px}} \approx \tfrac{H_{\text{m}}}{g},
\]
which define the search window for the Deep Template Matcher. We extract an oriented rectangular crop from the PCD and 3DGS zenith renders, as shown in Figure~\ref{fig:blocks_wrivinder}, to form a clean, scene-specific template for alignment in the DTM stage.

\subsection{Deep Template Matcher (DTM)}
\label{sec:dtm}

Once the zenith-view rendering is generated, the next task is to locate its corresponding region on the satellite image. Using the metric footprint estimated in Sec.~\ref{sec:metric}, we restrict the search to a bounded window on the satellite tile. Even within this region, cross-modal matching remains challenging due to the large viewpoint gap and the appearance differences between 3DGS renders and satellite imagery.

We evaluated several off-the-shelf cross-view and cross-modal matchers (e.g., RoMA~\cite{edstedt2024roma} and MatchAnything-LoFTR/RoMA~\cite{he2025matchanything}), but none produced reliable correspondences for this setting. We therefore adopt a test-time \textit{Deep Template Matcher (DTM)}: a lightweight Siamese CNN with a ResNet-18 backbone~\cite{he2016deep} that compares the zenith template with candidate satellite crops and outputs a similarity score. This provides a simple and efficient way to measure alignment quality without requiring any paired ground–satellite training data.

\input{sec/5__exps_table}

\subsubsection{Pseudo Ground Truth Data Generation}
To enable self-supervised optimization, we generate pseudo ground-truth patch pairs directly from the satellite image. Each pair consists of two crops whose dimensions match the estimated metric footprint of the zenith render. To approximate the appearance of a 3DGS render, the second crop is augmented with Gaussian blur and localized intensity perturbations (“blobby jitter”). These synthetic pairs provide pseudo-aligned supervision for learning viewpoint- and modality-invariant similarity.

\subsubsection{DTM Training in Self-Supervised Fashion}
The ResNet-18 Siamese model predicts the IoU between two input crops. During training, we sample many crop pairs from the satellite image and supervise the network with their true IoU values, encouraging it to identify when two regions correspond despite mild appearance variations.  
At inference time, the 3DGS zenith crop is fed into one branch of the network, while the other branch evaluates all candidate crops inside the satellite search window. The resulting similarity scores form a heatmap, and the peak response is selected as the zenith–satellite alignment. In practice, this lightweight matcher proves reliable across diverse scenes.

\subsection{Gaussian Splat and Camera Geolocator}
\label{sec:geolocator}
After localizing the zenith crop, we extract a slightly larger satellite patch around the predicted location and match it to the 3DGS zenith render using a cross-modal point matcher such as \textit{MatchAnything-RoMA}. Restricting the matcher to this localized region yields far more reliable correspondences than attempting global matching on the full satellite image.

These correspondences assign latitude and longitude to pixels in the 3DGS zenith render, which are then back-projected to the 3DGS points. Because the SfM and 3DGS reconstructions share a common coordinate system, each SfM point inherits geographic coordinates from its nearest 3DGS neighbor. A RANSAC-based similarity transform aligns the SfM reconstruction to these world coordinates, producing GPS estimates for all ground cameras. This completes the pipeline: starting from only ground images and a satellite map, Wrivinder recovers geographically aligned camera poses.

%% file: sec/5__exps_table.tex
\renewcommand{\arraystretch}{1.25}  % increases row height by 25%

\begin{table*}[!htbp]
\resizebox{\textwidth}{!}{%
\begin{tabular}{|c|c|c|c|c|c|c|ccc|}
\hline
\rowcolor[HTML]{F9CB9C} 
\cellcolor[HTML]{F9CB9C}                                     & \cellcolor[HTML]{F9CB9C}                               & \cellcolor[HTML]{F9CB9C}                                 & \cellcolor[HTML]{F9CB9C}                                                                              & \cellcolor[HTML]{F9CB9C}                                                                         & \cellcolor[HTML]{F9CB9C}                                                                                & \cellcolor[HTML]{F9CB9C}                                                                                                  & \multicolumn{3}{c|}{\cellcolor[HTML]{F9CB9C}Geolocation Error (in meters)}                                                                                                                                                                                                                                         \\ \cline{8-10} 
\rowcolor[HTML]{F9CB9C} 
\multirow{-2}{*}{\cellcolor[HTML]{F9CB9C}MC-Sat Scene Name} & \multirow{-2}{*}{\cellcolor[HTML]{F9CB9C}Dataset Type} & \multirow{-2}{*}{\cellcolor[HTML]{F9CB9C}Source Dataset} & \multirow{-2}{*}{\cellcolor[HTML]{F9CB9C}\begin{tabular}[c]{@{}c@{}}Satellite \\ Source\end{tabular}} & \multirow{-2}{*}{\cellcolor[HTML]{F9CB9C}\begin{tabular}[c]{@{}c@{}}Image \\ Count\end{tabular}} & \multirow{-2}{*}{\cellcolor[HTML]{F9CB9C}\begin{tabular}[c]{@{}c@{}}Run Time \\ (in mins)\end{tabular}} & \multirow{-2}{*}{\cellcolor[HTML]{F9CB9C}\begin{tabular}[c]{@{}c@{}}World2Model \\ RMSE\end{tabular}} & \multicolumn{1}{c|}{\cellcolor[HTML]{F9CB9C}\begin{tabular}[c]{@{}c@{}}Geolocation RMSE\\ (67th Percentile)\end{tabular}} & \multicolumn{1}{c|}{\cellcolor[HTML]{F9CB9C}\begin{tabular}[c]{@{}c@{}}Geolocation RMSE\\ (Mean)\end{tabular}} & \begin{tabular}[c]{@{}c@{}}Geolocation \\ Centroid Error\end{tabular} \\ \hline
APL Front Door                                               & Image Density                                          & ULTRAA                                                   & NAIP                                                                                                  & 100                                                                                              & 228                                                                                                     & 0.96                                                                                                                      & \multicolumn{1}{c|}{1.86}                                                                                                 & \multicolumn{1}{c|}{1.96}                                                                                      & 0.86                                                                  \\ \hline
APL Back Door                                                & Image Density                                          & ULTRAA                                                   & NAIP                                                                                                  & 100                                                                                              & 296                                                                                                     & 1.13                                                                                                                      & \multicolumn{1}{c|}{2.56}                                                                                                 & \multicolumn{1}{c|}{2.82}                                                                                      & 0.76                                                                  \\ \hline
MUTC A09                                                     & Reconstructed Area                                     & ULTRAA                                                   & ESRI                                                                                                  & 334                                                                                              & 484                                                                                                     & 3.36                                                                                                                      & \multicolumn{1}{c|}{18.33}                                                                                                & \multicolumn{1}{c|}{18.86}                                                                                     & 17.34                                                                 \\ \hline
MUTC A10                                                     & Reconstructed Area                                     & ULTRAA                                                   & ESRI                                                                                                  & 271                                                                                              & 522                                                                                                     & 15.76                                                                                                                     & \multicolumn{1}{c|}{17.59}                                                                                                & \multicolumn{1}{c|}{17.82}                                                                                     & 16.96                                                                 \\ \hline
siteSTR0001 (South America)                                  & Reconstructed Area                                     & VisymScenes                                              & ESRI                                                                                                  & 2705                                                                                             & 1560                                                                                                    & NaN                                                                                                                       & \multicolumn{1}{c|}{56.88}                                                                                                & \multicolumn{1}{c|}{57.22}                                                                                     & 43.82                                                                 \\ \hline
siteSTR0003 (South America)                                  & Reconstructed Area                                     & VisymScenes                                              & ESRI                                                                                                  & 2645                                                                                             & 2170                                                                                                     & NaN                                                                                                                       & \multicolumn{1}{c|}{15.22}                                                                                                & \multicolumn{1}{c|}{17.67}                                                                                     & 11.56                                                                 \\ \hline
siteSTR0007 (South America)                                  & Reconstructed Area                                     & VisymScenes                                              & ESRI                                                                                                  & 2619                                                                                             & 1485                                                                                                    & 16.59                                                                                                                     & \multicolumn{1}{c|}{32.96}                                                                                                & \multicolumn{1}{c|}{33.12}                                                                                     & 24.18                                                                 \\ \hline
siteSTR0008 (South America)                                  & Reconstructed Area                                     & VisymScenes                                              & ESRI                                                                                                  & 2652                                                                                             & 1805                                                                             & NaN                                                                                               & \multicolumn{1}{c|}{73.58}                                                                          & \multicolumn{1}{c|}{86.44}                                                               & 72.39                                           \\ \hline
siteSTR0098 (Univ. Philippines)                              & Reconstructed Area                                     & VisymScenes                                              & ESRI                                                                                                  & 2805                                                                                             & 1950                                                                                                    & NaN                                                                                                                       & \multicolumn{1}{c|}{16.55}                                                                                                & \multicolumn{1}{c|}{18.32}                                                                                     & 6.24                                                                  \\ \hline
AMES Hall                                                    & Reconstructed Area                                     & AMES                                                     & ESRI                                                                                                  & 1605                                                                                             & 1055                                                                             & 3.52                                                                                               & \multicolumn{1}{c|}{56.84}                                                                          & \multicolumn{1}{c|}{59.17}                                                               & 55.42                                           \\ \hline
% siteSTR0057 (US)                                             & Reconstructed Area                                     & VisymScenes                                              & ESRI                                                                                                  & 776                                                                                              & \cellcolor[HTML]{F4CCCC}TBD                                                                             & \cellcolor[HTML]{F4CCCC}TBD                                                                                               & \multicolumn{1}{c|}{\cellcolor[HTML]{F4CCCC}TBD}                                                                          & \multicolumn{1}{c|}{\cellcolor[HTML]{F4CCCC}TBD}                                                               & \cellcolor[HTML]{F4CCCC}TBD                                           \\ \hline
siteSTR0058 (US)                                             & Reconstructed Area                                     & VisymScenes                                              & ESRI                                                                                                  & 654                                                                                              & 885                                                                                                     & 3.22                                                                                                                      & \multicolumn{1}{c|}{11.23}                                                                                                & \multicolumn{1}{c|}{11.88}                                                                                     & 10.55                                                                 \\ \hline
siteSTR0059 (US)                                             & Image Density                                          & VisymScenes                                              & ESRI                                                                                                  & 85                                                                                               & 272                                                                                                     & 0.86                                                                                                                      & \multicolumn{1}{c|}{32.55}                                                                                                & \multicolumn{1}{c|}{32.13}                                                                                     & 31.86                                                                 \\ \hline
siteACC0003-finearts\_Top\_Right                             & Image Density                                          & ACC-NVS                                                  & ESRI                                                                                                  & 277                                                                                              & 425                                                                                                     & 4.66                                                                                                                      & \multicolumn{1}{c|}{2.86}                                                                                                 & \multicolumn{1}{c|}{3.02}                                                                                      & 2.16                                                                  \\ \hline
% siteACC0004-mill19\_Left\_Side                               & Image Density                                          & ACC-NVS                                                  & ESRI                                                                                                  & 313                                                                                              & \cellcolor[HTML]{F4CCCC}TBD                                                                             & \cellcolor[HTML]{F4CCCC}TBD                                                                                               & \multicolumn{1}{c|}{\cellcolor[HTML]{F4CCCC}TBD}                                                                          & \multicolumn{1}{c|}{\cellcolor[HTML]{F4CCCC}TBD}                                                               & \cellcolor[HTML]{F4CCCC}TBD                                           \\ \hline
siteACC0004-mill19\_Right\_Side                              & Image Density                                          & ACC-NVS                                                  & ESRI                                                                                                  & 732                                                                                              & 650                                                                                                     & 1.16                                                                                                                      & \multicolumn{1}{c|}{62.53}                                                                                                & \multicolumn{1}{c|}{64.86}                                                                                     & 44.9                                                                  \\ \hline
% siteACC0152-park-plaza\_Front\_Door                          & Image Density                                          & ACC-NVS                                                  & ESRI                                                                                                  & 1388                                                                                             & \cellcolor[HTML]{F4CCCC}TBD                                                                             & \cellcolor[HTML]{F4CCCC}TBD                                                                                               & \multicolumn{1}{c|}{\cellcolor[HTML]{F4CCCC}TBD}                                                                          & \multicolumn{1}{c|}{\cellcolor[HTML]{F4CCCC}TBD}                                                               & \cellcolor[HTML]{F4CCCC}TBD                                           \\ \hline
siteACC0153-rec-center\_Front\_Door                          & Image Density                                          & ACC-NVS                                                  & ESRI                                                                                                  & 915                                                                                              & 745                                                                                                     & 1.24                                                                                                                      & \multicolumn{1}{c|}{56.22}                                                                                                & \multicolumn{1}{c|}{59.15}                                                                                     & 51.33                                                                 \\ \hline
\end{tabular}%
}
\caption{Quantitative results on the MC-Sat benchmark.
We report the number of images, run time, SfM alignment quality (World2Model RMSE), and final geo-localization error across all evaluated scenes. Lower values indicate better alignment.}
\label{tab:quant_res}
\end{table*}

\renewcommand{\arraystretch}{1.0}  % increases row height by 25%

%% file: sec/exp_portrait.tex
\begin{figure}[!htbp]
    \centering
    \includegraphics[width=1.0\columnwidth]{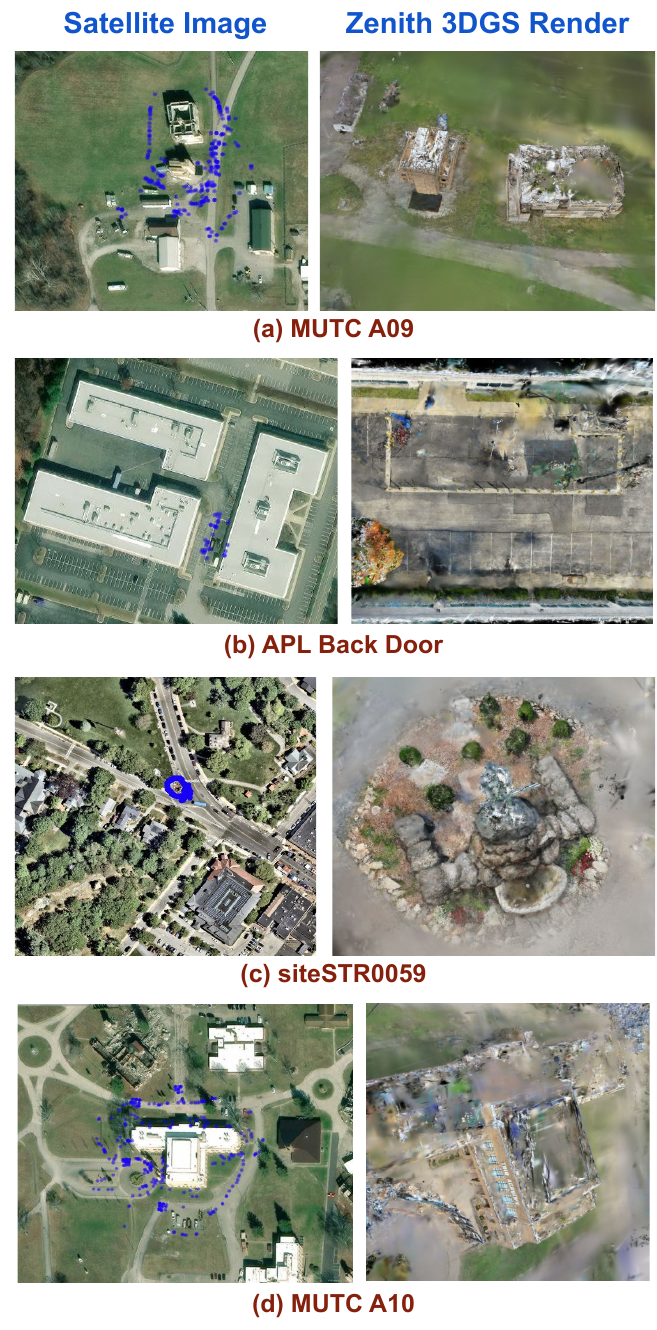}
    \vspace{-0.3cm}
    \caption{
Satellite–render pairs for several MC-Sat scenes.  
In each case, the left image shows the satellite view (with blue dots indicating the ground-truth camera locations) and the right image shows the corresponding 3DGS zenith rendering produced by Wrivinder.  
Gaps, blurring, and missing structures in the reconstruction often make alignment more ambiguous and are a primary source of the higher errors observed in some scenes.
}
\vspace{-0.6cm}
    \label{fig:zenith_renderings}
    % \vspace{-0.5cm}
\end{figure}

% \newpage
\section{Experiments}
\label{sec:exp}
 
\subsection{Implementation Details}

We evaluate Wrivinder on the MC-Sat dataset introduced in Sec.~\ref{sec:dataset}. All experiments use the \textit{HLOC} pipeline with \textit{PyCOLMAP} as the SfM backend. The scene graph is initialized with a combination of \textit{NetVLAD} and \textit{EigenPlaces} descriptors before geometric verification. Among the SfM variants we tested, this configuration produced the most stable reconstructions across MC-Sat’s diverse outdoor scenes.

For the dense 3D representation, we use Octree-GS~\cite{ren2024octree} to construct a 3D Gaussian Splatting model, which is subsequently rendered from the estimated zenith viewpoint for satellite alignment. Unless otherwise noted, all other components follow the settings described in Sec.~\ref{sec:method}. Localization accuracy is reported using the metrics defined below.

\subsection{Metrics}

We first evaluate the quality of the underlying SfM reconstruction.  
\textbf{World2Model SfM RMSE} measures how well the SfM camera centers align to the satellite frame.  
A similarity transform is estimated using the triplet-based Procrustes aligner~\cite{11084709}, and we report the 67th percentile alignment error.  
This reflects the stability of the SfM solution—poor alignment typically propagates to weaker 3DGS rendering and degraded geo-localization.  
If fewer than 67\% of images register into the dominant SfM cluster, this metric is reported as NaN.

To assess final camera geo-localization, we report three metrics:

\begin{itemize}
    \item \textbf{Mean Geolocation RMSE:} mean haversine distance between predicted and ground-truth camera coordinates.
    \item \textbf{67th Percentile Geolocation RMSE:} provides a robust measure less sensitive to outliers.
    \item \textbf{Geolocation Centroid Error:} Haversine distance between predicted and ground-truth camera centroids, capturing large-scale drift.
\end{itemize}

These metrics form an initial benchmark for ground-to-satellite alignment. Future versions of MC-Sat will include additional scenes and richer annotations, enabling evaluations such as IoU-based template-matching accuracy and comparisons across multiple satellite resolutions.
We also report \textbf{per-scene run time} (in minutes) to characterize computational cost.

\subsection{Results and Discussion}

Table~\ref{tab:quant_res} summarizes quantitative results across all MC-Sat scenes. Run time scales roughly linearly with the number of input images, with the SfM stage dominating the computational cost.

Wrivinder achieves accurate geolocation on several scenes, particularly those with dense coverage or compact spatial layouts.  Performance decreases on large \textit{Reconstructed Area} scenes, where many surfaces—especially rooftops and elevated structures—are never observed from the ground. As shown in Fig.~\ref{fig:zenith_renderings}, these unobserved regions lead to gaps in the zenith render, making alignment more challenging and reducing template-matching reliability.  Scenes with higher World2Model SfM RMSE exhibit this trend most clearly, indicating a strong dependency on reconstruction stability.

Some variability in output is expected, as several components—most notably SfM—use RANSAC-based procedures. Nevertheless, Wrivinder provides a unified zero-shot framework for geometry-driven ground-to-satellite alignment and establishes a first baseline on MC-Sat. The results reveal both the feasibility of this approach and key opportunities for improvement.

A promising direction is to incorporate semantic cues directly into the 3DGS representation.  
Recent advances in semantic Gaussians~\cite{guo2024semantic} suggests that semantically enriched splats could reduce artifacts in zenith renders and provide more stable cues for alignment.

Additional qualitative and quantitative results are provided in the supplementary material.

\vspace{-0.35cm}
\section{Conclusion}
\vspace{-0.2cm}

We presented \textbf{Wrivinder}, a zero-shot, geometry-driven framework for aligning ground imagery to geo-registered satellite maps, and introduced \textbf{MC-Sat}, the first dataset that links multi-view ground captures, 3D reconstructions, and overhead imagery across diverse outdoor environments. Together, they establish a unified setting and baseline for studying metrically evaluated ground-to-satellite alignment beyond road-centric or paired-image benchmarks.

Our experiments demonstrate that geometry-centered aggregation—combining SfM, 3D Gaussian Splatting, semantic cues, and test-time alignment—can recover meaningful geolocation across challenging real-world scenes. MC-Sat also reveals where future progress is most needed, particularly in handling unobserved surfaces and improving stability in large, complex environments.

We view Wrivinder and MC-Sat as a foundation for new approaches that integrate richer semantics, more robust cross-view matching, and learned geometric priors. By making both resources publicly available, we hope to encourage further research on cross-view spatial reasoning in settings where paired supervision is unrealistic or unavailable.

%% file: sec/acknowledgements.tex
\section{Acknowledgments}
% \vspace{-0.3cm}
This research is supported by the Intelligence Advanced Research Projects Activity (IARPA) via Department of Interior/ Interior Business Center (DOI/IBC) contract number 140D0423C0076. The U.S. Government is authorized to reproduce and distribute reprints for Governmental purposes notwithstanding any copyright annotation thereon. The views and conclusions contained herein are those of the authors and should not be interpreted as necessarily representing the official policies or endorsements, either expressed or implied, of IARPA, DOI/IBC, or the U.S. Government. We would like to thank Jason Bunk for insights and assistance during the initial phase of this project.